\documentclass[10pt,twocolumn,letterpaper]{article}

\usepackage{cvpr}              

\usepackage{multirow}
\usepackage{subcaption}
\usepackage{graphicx}
\usepackage{cuted}      
\usepackage{capt-of}    

\definecolor{cvprblue}{rgb}{0.21,0.49,0.74}
\usepackage[pagebackref,breaklinks,colorlinks,allcolors=cvprblue]{hyperref}

\def\paperID{5364} 
\def\confName{CVPR}
\def\confYear{2026}

\title{Image Generation as a Visual Planner for Robotic Manipulation}

\author{
Pang Ye\\
Southern China University of Technology\\
Guangzhou, China\\
{\tt\small 202264690373@mail.scut.edu.cn}
}
\begin{document}
\maketitle
\begin{strip}
    \centering
    \vspace{-0.8cm}  
    \includegraphics[width=\linewidth]{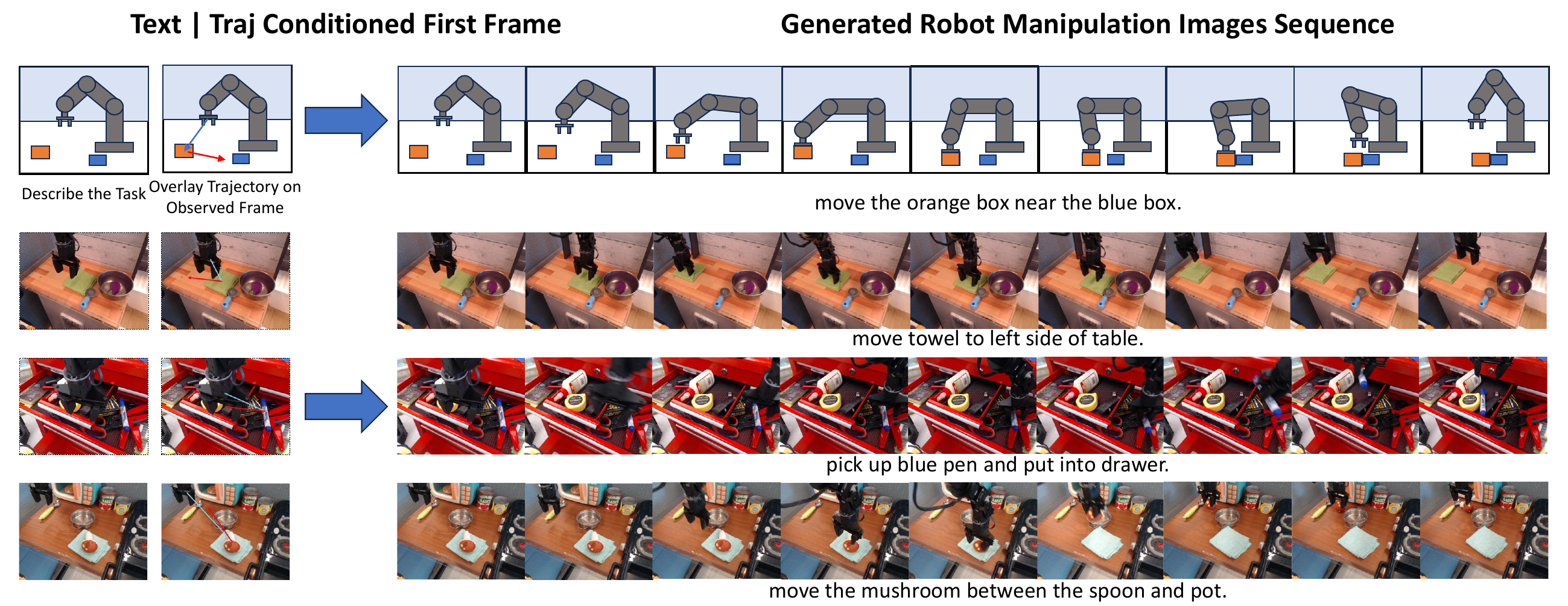} 
    \captionof{figure}{
        \textbf{Overview of our work.}
        We convert a pretrained image generator into a visual planner
        that synthesizes $3{\times}3$ manipulation grids under either
        text or trajectory conditioning.
    }
    \label{fig:teaser}
\end{strip}
\begin{abstract}
Generating realistic robotic manipulation videos is a key step toward unifying perception, planning, and action in embodied agents.
While existing video diffusion models require large domain-specific datasets and struggle to generalize, recent image generation models trained on language--image corpora exhibit strong compositionality—including the ability to synthesize temporally coherent \textit{grid images}.
This suggests a latent capacity for \textit{video-like generation} even without explicit temporal modeling.

We explore whether such models can serve as \textit{visual planners} for robots when lightly adapted via LoRA finetuning.
We propose a two-part framework with: (1) \textbf{text-conditioned generation}, using a language instruction and the first frame; and (2) \textbf{trajectory-conditioned generation}, using a 2D trajectory overlay and the same initial frame.
Experiments on \textbf{Jaco Play dataset}, \textbf{Bridge V2} and \textbf{RT1 dataset} show that both modes produce smooth, coherent robot videos aligned with their respective conditions.

Our findings indicate that pretrained image generators encode transferable temporal priors and can serve as \textit{video-like robotic planners} under minimal supervision.
Code is released at \href{https://github.com/pangye202264690373/Image-Generation-as-a-Visual-Planner-for-Robotic-Manipulation}{https://github.com/pangye202264690373/Image-Generation-as-a-Visual-Planner-for-Robotic-Manipulation}.
\end{abstract}    
\section{Introduction}
\begin{figure*}[t]
    \centering
    \includegraphics[width=\textwidth]{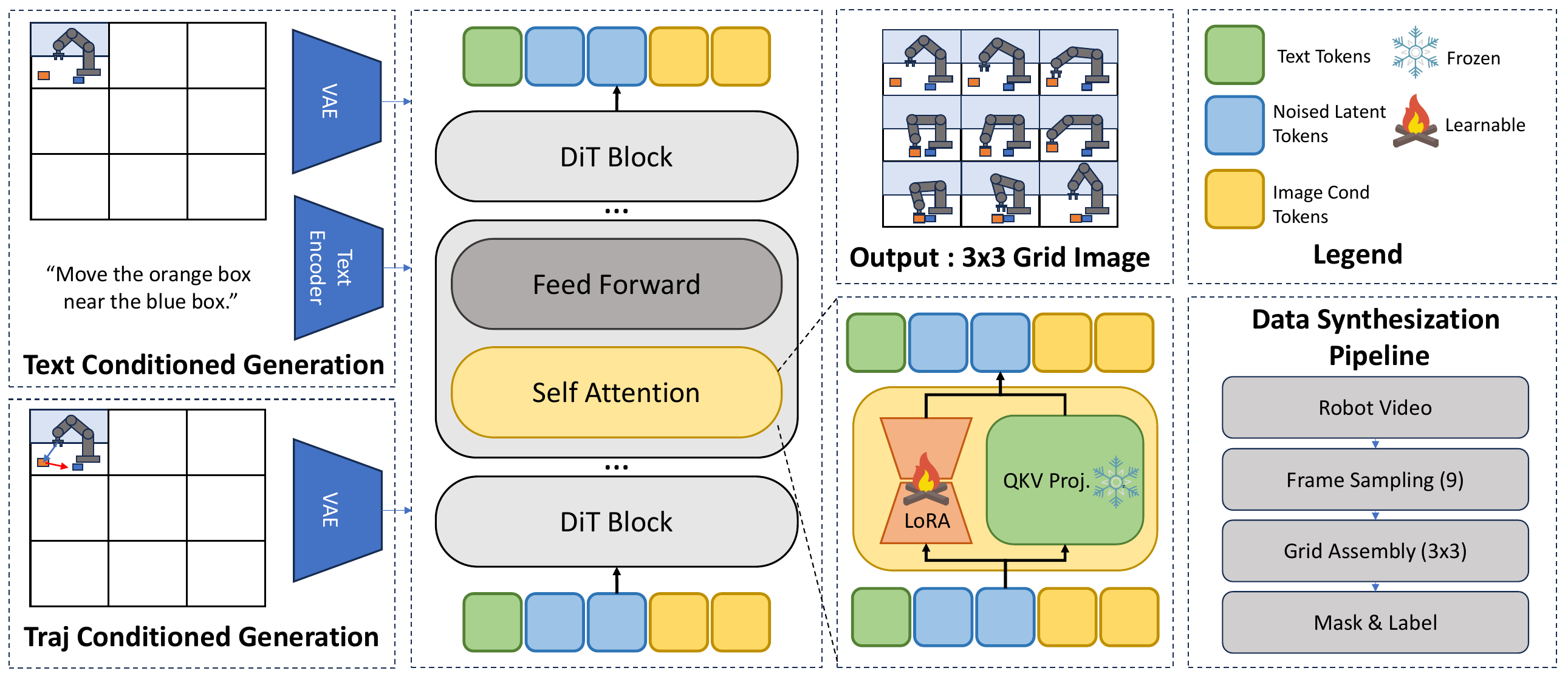}
    \caption{\textbf{Overall framework of our approach.}
    Our method adapts a pretrained image generator (DiT backbone with LoRA adapters)
    into a controllable video-like synthesizer that outputs a $3{\times}3$ grid image representing
    a short manipulation sequence.
    The upper branch shows the \textit{text-conditioned} generation, where a language instruction
    and the first observed frame are encoded by the Text Encoder and VAE respectively.
    The lower branch shows the \textit{trajectory-conditioned} generation, where a 2D path is
    rendered over the first frame and encoded similarly.
    Both branches share the same DiT architecture with LoRA applied to attention projections.
    On the right, the data synthesization pipeline illustrates how robot videos are processed:
    frame sampling, grid assembly, and masking for conditional supervision.
    On the right, the data synthesization pipeline illustrates how robot videos are processed:
    frame sampling, grid assembly, and masking for conditional supervision. See fig \ref{fig:data_synthesis_pipeline} for details.
    }
    \label{fig:overall_architecture}
\end{figure*}
\label{sec:intro}

Generating realistic \textit{robotic manipulation videos} is a fundamental challenge for embodied intelligence and robot planning.
A model capable of visually imagining how a robot executes a high-level instruction—e.g., ``pick up the red cup and place it on the table''—would bridge perception, planning, and action in a unified manner.
Such a capability enables robots to reason about future states and physical interactions directly in the visual domain.

However, learning to synthesize coherent robot videos remains difficult.
Existing video diffusion or action-conditioned models require large-scale domain-specific robot datasets and often fail to generalize beyond seen tasks or scenes.
Their heavy temporal modeling also leads to high computational cost and limited semantic understanding.

Meanwhile, modern \textit{image generation models} such as diffusion or transformer-based architectures have achieved remarkable progress in realism and semantic alignment after training on billions of language–image pairs.
Beyond single-frame synthesis, these models exhibit strong \textit{compositional generation} capabilities:
they can arrange multiple coherent images within a single grid layout, preserving spatial and semantic consistency across sub-images.
Interestingly, such grids often resemble step-by-step visual progressions—akin to short video clips—with implicit temporal transitions between neighboring cells.
This phenomenon reveals that image generators may already contain latent capacity for \textit{video-like generation}, even without explicit temporal modeling.
Such visual coherence, compositionality, and physical plausibility are precisely the ingredients required for robotic planning.

Thus, pretrained image generators already possess rich priors over objects, physics, and composition—qualities that are essential for robotic planning.

Motivated by these observations, we hypothesize that large-scale image generation models can act as \textit{visual planners} for robots, capable of synthesizing temporally coherent manipulation sequences from minimal conditional cues.
Instead of designing new video architectures, we explore how existing image generators can be adapted, via lightweight \textbf{LoRA}-based~\cite{hu2021lora} finetuning, to produce robot manipulation videos guided by simple conditions, as shown in figure~\ref{fig:teaser}. 

\noindent\textbf{Our exploration consists of two complementary parts:}

\begin{enumerate}[leftmargin=1.5em]
    \item \textbf{Text-Conditioned Generation (Text + First Frame).}  
    In the first setting, the model receives a natural-language instruction together with the first observed image of the robot and environment.
    Through LoRA adaptation, it learns to generate temporally coherent sequences that visually execute the instruction.
    This experiment studies the model’s ability to perform \emph{text-following}—understanding high-level semantics and transforming them into plausible motion imagery.

    \item \textbf{Trajectory-Conditioned Generation (Trajectory + First Frame).}  
    In the second setting, the model is conditioned on a rendered 2D end-effector trajectory over the initial frame, along with the same first image.
    The model learns to follow the provided spatial guidance while maintaining physical consistency, emphasizing \emph{trajectory-following}—precise, continuous control of robot motion through visual synthesis.
\end{enumerate}

Together, these two paradigms investigate complementary aspects of generative control: semantic reasoning through language versus spatial reasoning through motion trajectories.
They reveal how large-scale image generators can be repurposed as \textit{video-like robotic planners} when provided with appropriate conditioning signals.

\noindent\textbf{Experimental highlights.}
We evaluate both paradigms on multiple robotic video datasets, including \textbf{JacoPlay}, \textbf{Bridge}, and \textbf{RT1 datasets}.
Our adapted models produce smooth, semantically aligned motion sequences that follow their respective conditioning modalities.
The text-conditioned model generalizes to unseen instructions, while the trajectory-conditioned model excels at spatial precision.
Both outperform task-specific video diffusion baselines in perceptual quality and action fidelity 

\noindent\textbf{Contributions.}
We summarize our main contributions as follows:
\begin{itemize}
    \item \textbf{Image Generation as Visual planning.}
    We reveal that pretrained image generation models, known for producing coherent grid images, can be adapted to serve as \emph{video-like planners} for robotic manipulation.
    \item \textbf{Two-part Conditional Framework.}
    We introduce a lightweight LoRA-based adaptation with two complementary conditioning modes—\emph{text + first frame} and \emph{trajectory + first frame}—enabling both semantic and spatial control.
    \item \textbf{Comprehensive Evaluation and Insights.}
    Extensive experiments show strong text-following and trajectory-following performance across datasets, highlighting how these conditioning forms jointly enable visual reasoning and motion synthesis.
\end{itemize}

Our framework builds on a standard diffusion/DiT backbone with LoRA adapters for conditioning, trained under first-frame supervision and autoregressive decoding.

\section{Related Work}

\paragraph{Image Generation Models.}
Diffusion models have become the de facto approach for high-fidelity image synthesis. Diffusion Transformers (DiT) replace the U-Net backbone with a Transformer and set strong image generation performance on ImageNet, cementing the ``Transformer-ization'' of diffusion backbones \cite{peebles2023dit}. For controllability, ControlNet adds a zero-initialized control branch on top of a frozen text-to-image backbone to inject spatial hints such as edges, depth, or human pose, substantially improving composition and layout control \cite{zhang2023controlnet}. While these advances establish powerful \emph{static} image controllability, bridging from compositional control in single images to \emph{temporally consistent} and \emph{task-aligned} control in videos remains challenging—motivating us to adapt DiT-style models to robotic manipulation videos.

\paragraph{Video Generation and Diffusion.}
Recent text-to-video (T2V) progress shows rapid improvements in temporal coherence and long-range consistency \cite{sora, gen3, genie3, minimax, veo3, moviegen, hunyuanvideo, i2vgenxl, pyramidflow, ma2024followyouremoji, ma2025controllable, ma2025followcreation, ma2025followfaster, ma2025followyourclick, ma2025followyourmotion}. \emph{Lumiere} proposes a single space–time U-Net that generates the full video in one pass, improving cross-frame consistency compared to keyframe-then-upsampling pipelines \cite{bartal2024lumiere}. OpenAI’s \emph{Sora} demonstrates sustained object persistence and long-horizon dynamics under text prompts, reinforcing the emerging view of video generators as ``world simulators'' \cite{openai2024sora}. Surveys synthesize the design space and challenges of diffusion-based video generation, including temporal modeling and evaluation \cite{melnik2024_videodiff_survey}. Yet generic T2V models still struggle with \emph{precise} motion and interaction constraints. To address this, trajectory-/tracking-conditioned video diffusion has emerged: TrackDiffusion conditions on multi-object tracklets for instance-consistent motion control \cite{li2025trackdiffusion}, while Motion Prompting conditions on sparse/dense point trajectories to jointly control object and camera motion \cite{geng2025motionprompting}. These trends suggest that explicit spatiotemporal constraints are key to pushing video diffusion from ``compelling visuals'' to \emph{task-usable} generations—exactly the regime required for robot manipulation videos.


\paragraph{Visual Imitation and Robotic Video Synthesis.}
A growing line of work closes the loop between generative video models and robot learning \cite{rt1, roboflamingo, leo, gr1, octo, openvla, tinyvla, pi0, uniact, openvla, gr00t}. \emph{RIGVid} uses AI-generated task videos (filtered by a VLM), estimates 6-DoF trajectories from the videos, and executes the resulting rollouts on real robots—achieving complex tasks such as pouring and wiping without physical demonstrations \cite{patel2025rigvid}. \emph{Gen2Act} generates human execution videos in novel scenes and conditions a \emph{single} policy on those videos to generalize manipulation to unseen objects and motions \cite{bharadhwaj2025gen2act}. Large cross-domain datasets such as BridgeData~V2 provide broad task and environment coverage for scalable imitation learning \cite{walke2023bridgedatav2}. Compared to methods that rely on heavy end-to-end T2V or expensive data collection, we take a complementary path: starting from a powerful \emph{image} diffusion backbone (DiT), we add parameter-efficient LoRA adapters and explicit text/trajectory conditions to synthesize \emph{robotic manipulation videos} with fine-grained, repeatable motion control, striking a favorable balance between cost, controllability, and generality.

\paragraph{Controllable Generation.}
Early controllable generation in diffusion exploits \emph{explicit conditioning branches} or \emph{guided objectives} to steer pretrained models. A seminal instance is \textbf{ControlNet}, which freezes a text-to-image backbone and adds a zero-initialized control branch to inject spatial conditions (edges, depth, pose), enabling precise composition without degrading the base model \cite{zhang2023controlnet, zhang2024ssr, song2024processpainter, zhang2024stable, zhang2025stable}. More recently, surveys have systematized controllability along conditioning types (layout/pose/depth/boxes), guidance strategies, and training vs.\ training-free control for both image and video diffusion \cite{chen2025transanimate}, highlighting open challenges in multi-condition fusion and temporally consistent control \cite{cao2024_controllable_t2i_survey}. As representative applications, \textbf{GLIGEN} introduces open-set \emph{grounded} generation by injecting region-level grounding (e.g., bounding boxes) via gated adapters while keeping the base diffusion model frozen \cite{li2023gligen}; on the video side, \textbf{Motion Prompting} conditions generation on sparse or dense \emph{point trajectories}, jointly controlling object and camera motion for fine-grained, temporally coherent synthesis \cite{geng2025motionprompting}. These advances motivate our design to combine LoRA-based parameter-efficient adapters with text/trajectory conditions to achieve repeatable, robot-centric controllability in manipulation videos. Recently, transformer-based diffusion models have enabled semantic and spatial conditions to be represented uniformly as token sequences. These condition tokens are incorporated into the generative process via multimodal attention or concatenation strategies, as seen in DiT-style frameworks such as EasyControl\cite{zhang2025easycontrol}. This token-centric perspective has significantly pushed forward conditional image generation~\cite{wan2024grid, song2025makeanything, song2025layertracer, huang2025photodoodle, guo2025any2anytryon, jiang2025personalized, wang2025diffdecompose, gong2025relationadapter}, offering better scalability, cleaner architectural design, and more efficient support for diverse or high-resolution conditioning inputs.

\section{Method}


\begin{figure*}[htb]
    \centering
    \includegraphics[width=\textwidth]{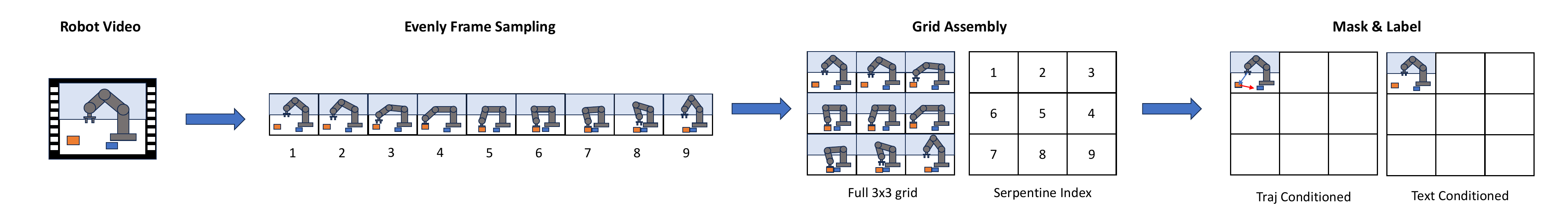}
    \caption{\textbf{Data Synthesis Pipeline.}
    From each robot video, nine frames are uniformly sampled and arranged
    into a $3{\times}3$ grid following a serpentine temporal order 
    ($1{\rightarrow}2{\rightarrow}3$, $6{\leftarrow}5{\leftarrow}4$, $7{\rightarrow}8{\rightarrow}9$).
    Only the top-left cell remains visible as the conditioning frame, while the other cells are masked to zero.
    For the trajectory-conditioned variant, a 2D end-effector path is overlaid on the first frame
    (red$\rightarrow$blue indicating temporal progression).
    The resulting masked grid serves as the model input,
    and the complete grid as the reconstruction target for supervised training.}
    \label{fig:data_synthesis_pipeline}
\end{figure*}

Our goal is to transform a pretrained image generation model into a controllable
\textit{video-like} synthesizer capable of producing short robotic manipulation sequences
from minimal conditional inputs. The overall method comprises three major components:
(i) an architectural adaptation that integrates LoRA modules into a diffusion transformer backbone for efficient finetuning;
(ii) dual conditioning strategies—text-based and trajectory-based—that enable semantic or spatial control of the generated motion;
and (iii) a data synthesis pipeline that converts real robot videos into supervision-ready grid–mask pairs.
Figure~\ref{fig:overall_architecture} provides an overview of this framework,
and Figure~\ref{fig:data_synthesis_pipeline} details the data construction process.


\subsection{Images Sequence Framework}

\paragraph{Backbone and Latent Space.}
We build on \textbf{FLUX.1-dev}, a large rectified-flow/transformer generator from Black Forest Labs, deployed via its official implementation and docs.
Following latent diffusion/transformer practice, images are encoded to a latent tensor $z=\mathcal{E}(x)$ by a VAE encoder $\mathcal{E}$ and decoded by $\mathcal{D}$. The generator $\mathcal{G}_\theta$ operates on latents and is conditioned by language and/or a rendered trajectory.

\paragraph{Inputs and Outputs.}
Let $\{I_t\}_{t=1}^9$ denote nine temporally ordered frames, 
each represented as a matrix $D^{\text{img}_t}\!\in\!\mathbb{R}^{m\times n}$.
We construct a $3\times3$ block matrix $\mathbf{D}$:
\begin{equation}
\mathbf{D} =
\begin{bmatrix}
D^{\text{img}_1} & D^{\text{img}_2} & D^{\text{img}_3}\\
D^{\text{img}_6} & D^{\text{img}_5} & D^{\text{img}_4}\\
D^{\text{img}_7} & D^{\text{img}_8} & D^{\text{img}_9}
\end{bmatrix},
\label{eq:grid_matrix}
\begin{aligned}
\mathbf{D} &\in \mathbb{R}^{3m\times 3n},\\
D^{\text{img}_t} &\in \mathbb{R}^{m\times n}.
\end{aligned}
\end{equation}

This spatial arrangement keeps temporally adjacent frames close on the grid,
facilitating short-range temporal reasoning via local attention 
(see Sec.~\ref{sec:data}).

During training, we feed the model a \emph{partially observed} input grid 
$\tilde{\mathbf{D}}$, in which only the top-left block contains the 
observed image $D^{\text{img}_1}$, while all other regions are filled 
with zeros (black pixels):
\begin{equation}
\tilde{\mathbf{D}} =
\begin{bmatrix}
D^{\text{img}_1} & \mathbf{0} & \mathbf{0}\\
\mathbf{0} & \mathbf{0} & \mathbf{0}\\
\mathbf{0} & \mathbf{0} & \mathbf{0}
\end{bmatrix},
\label{eq:block_diag_matrix}
\end{equation}

Let $\mathcal{E}$ and $\mathcal{D}$ denote the VAE encoder and decoder,
and $\mathcal{G}_{\theta}$ the generator conditioned on $c$.
The network predicts the complete $3\times3$ grid as

\begin{equation}
\hat{\mathbf{D}} =
\mathcal{D}\!\left(
    \mathcal{G}_{\theta}\big(
        \mathcal{E}(\tilde{\mathbf{D}});\, c
    \big)
\right),
\label{eq:decoder_generation}
\end{equation}

\paragraph{LoRA.}
We adopt the Low-Rank Adaptation (LoRA) technique~\cite{hu2021lora} to finetune the frozen backbone generator $\mathcal G_\theta$ with a controlled parameter budget. Specifically, for each selected dense projection matrix $W\in\mathbb R^{d\times d}$ in the transformer blocks, we reparameterize the update as

\begin{equation}
\begin{aligned}
W' &= W + \Delta W, \quad 
    \Delta W = \alpha\, A\, B^\top,\\
A &\in \mathbb{R}^{d\times r}, \quad 
B \in \mathbb{R}^{d\times r}, \quad 
r \ll d,
\end{aligned}
\label{eq:lora_update}
\end{equation}

so that only $A$ and $B$ are trained while $W$ remains frozen.  
In our setting, we apply the adapters to the query and value projections in the self-attention layers, as well as the feed-forward projections. This choice is motivated by empirical evidence (e.g., the original LoRA paper) that adaptation in these subspaces suffices for downstream tasks.  
By constraining adaptation to $O(r\,d)$ parameters rather than $O(d^{2})$, LoRA enables efficient specialization of the image generator to robot-video domains without incurring inference latency or retraining the full model.

\subsection{Overall Architecture}
The proposed system is designed to convert a pretrained image generator into a controllable robotic video synthesizer, and is composed of three core modules (see Fig.~\ref{fig:overall_architecture} for an overview).  
First, the \textit{Text-Conditioned Generation} module uses a natural-language instruction together with the first observed robot–environment frame to generate a full $3{\times}3$ image grid representing a manipulation sequence.  
Second, the \textit{Trajectory-Conditioned Generation} module accepts a rendered 2D end-effector trajectory over the first frame and synthesizes the corresponding grid, thereby enabling fine-grained spatial motion control.  
Third, the \textit{Data Synthesis Pipeline} module constructs the training supervision by extracting equi-spaced frames from real robot videos, arranging them into grid layouts, and partitioning the first frame as the visible condition.  
Together, this architecture enables both semantic control (via text) and spatial control (via trajectory) within a unified generative framework for robotic manipulation video synthesis.

\subsection{Conditioning Strategies}

In our framework, we support two separate conditioning modalities, each corresponding to a distinct experimental branch: semantic guidance via text, and spatial motion guidance via trajectory. Below we outline each mode and the common inference strategy.

\paragraph{Text-Conditioned Generation.}
We define a natural-language instruction \(t\) (dataset-derived prompt plus a fixed template indicating grid semantics and robot-manipulation context). We embed \(t\) using both a CLIP encoder and a T5 tokenizer:
\begin{equation}
e_{\mathrm{clip}} = \mathrm{CLIP}(t)_{\mathrm{pool}}, \quad
E_{\mathrm{t5}} = \mathrm{T5}(t) \in \mathbb{R}^{L\times d},
\label{eq:text_encoders}
\end{equation}

These embeddings are duplicated per generation and injected as conditioning \(c_{\mathrm{text}} = \{e_{\mathrm{clip}}, E_{\mathrm{t5}}\}\) into the generator \(\mathcal G_\theta\) via cross-attention. Given the masked input grid \(\tilde X\), the generator produces
\begin{equation}
\hat{X} =
\mathcal{G}_{\theta}\bigl(
    \mathcal{E}(\tilde{X});\, c_{\mathrm{text}}
\bigr),
\label{eq:gen_forward}
\end{equation}

and the training follows the latent MSE objective.

\paragraph{Trajectory-Conditioned Generation.}
In the second branch, we condition the model on a 2D end-effector trajectory \(\tau = \{(x_s,y_s)\}_{s=1}^S\) rendered over the first frame \(I_1\). We produce the overlay image \(I_1^\tau = \mathrm{Overlay}(I_1,\tau)\), and build the masked grid:
\begin{equation}
\tilde{X}^\tau =
M \odot \Pi\bigl(I_1^\tau,\, \mathbf{0},\, \dots,\, \mathbf{0}\bigr),
\label{eq:masked_grid}
\end{equation}

where only the top-left cell is non-zero. The generator then predicts
\begin{equation}
\hat{X} =
\mathcal{G}_{\theta}\bigl(
    \mathcal{E}(\tilde{X}^\tau);\, c_{\varnothing}
\bigr),
\label{eq:gen_traj}
\end{equation}

This branch emphasizes spatial-motion control while preserving scene semantics from \(I_1\).

\paragraph{Single-Shot Grid Generation.}
Rather than performing autoregressive frame generation, both branches employ a single-shot synthesis strategy: the full \(3\times3\) grid is generated at once. This decision takes advantage of the strong compositional priors of image generation models and aligns with observations that such models can implicitly model short temporal sequences via grid layouts.

\subsection{Data Synthesis Pipeline}

\label{sec:data}

We construct the training supervision from TFDS-stored robot video sequences, 
including the \textbf{JacoPlay}, \textbf{BridgeV2}, and \textbf{RT-1} datasets. Fig \ref{fig:data_synthesis_pipeline}
Each episode provides a temporally ordered sequence of RGB frames $\{I_t\}_{t=1}^{T}$.
We uniformly sample $9$ frames along the trajectory to form a compact visual summary 
of the manipulation episode:
\begin{equation}
\{I_t\}_{t=1}^{9} =
\mathrm{SampleUniform}\bigl(\{I_t\}_{t=1}^{T},\, 9\bigr),
\label{eq:uniform_sampling}
\end{equation}

\paragraph{Grid assembly.}
The sampled frames are arranged into a spatial $3{\times}3$ grid image $\mathbf{D}\!\in\!\mathbb{R}^{3H\times3W\times3}$
through a placement operator $\Pi$:
\begin{equation}
\mathbf{D} =
\begin{bmatrix}
D^{\text{img}_1} & D^{\text{img}_2} & D^{\text{img}_3} \\
D^{\text{img}_6} & D^{\text{img}_5} & D^{\text{img}_4} \\
D^{\text{img}_7} & D^{\text{img}_8} & D^{\text{img}_9}
\end{bmatrix},
\quad
D^{\text{img}_k} \in \mathbb{R}^{H \times W \times 3},
\label{eq:grid_composition}
\end{equation}

This \emph{serpentine ordering} ($1{\rightarrow}2{\rightarrow}3$, $6{\leftarrow}5{\leftarrow}4$, $7{\rightarrow}8{\rightarrow}9$)
places temporally adjacent frames in close spatial proximity,
helping the transformer’s local attention capture short-range temporal dependencies
through neighboring spatial regions.

\paragraph{Masked Input for Conditional generation.}
During training, we feed the model a \emph{partially observed} input grid
$\tilde{\mathbf{D}}$, in which only the top-left block contains the first observed frame,
and all other regions are masked to zero (black pixels):
\begin{equation}
\tilde{\mathbf{D}} = \mathbf{M} \odot \mathbf{D},
\quad
\mathbf{M}(x,y) =
\begin{cases}
1, & \text{if } (x,y) \in \mathrm{block}\!\big(D^{\text{img}_1}\big), \\[2pt]
0, & \text{otherwise.}
\end{cases}
\label{eq:masked_conditioning}
\end{equation}

This masked grid provides the conditioning signal for the generator.

\paragraph{Trajectory Overlay.}
For the trajectory-conditioned branch, we render the 2D end-effector path 
$\tau=\{(x_s, y_s)\}_{s=1}^S$ directly onto the first frame.
Earlier segments are colored \textcolor{blue}{blue}, and later ones \textcolor{red}{red},
to indicate temporal progression:
\begin{equation}
I_1^{\tau} =
\mathrm{Overlay}\bigl(I_1,\, \tau_{\text{red}\rightarrow\text{blue}}\bigr),
\label{eq:overlay}
\end{equation}

The overlaid image $I_1^{\tau}$ replaces $D^{\text{img}_1}$ in the masked grid,
yielding $\tilde{\mathbf{D}}^{\tau}$.

\paragraph{Supervision Construction.}
The model is trained to reconstruct the complete ground-truth grid $\mathbf{D}_{gt}$ from the masked condition:
\begin{equation}
\mathcal{L}_{\text{lat}} =
\big\|
    \mathcal{E}(\mathbf{D}_{gt}) -
    \mathcal{E}(\hat{\mathbf{D}})
\big\|_2^2,
\quad
\hat{\mathbf{D}} =
\mathcal{D}\!\left(
    \mathcal{G}_\theta\big(
        \mathcal{E}(\tilde{\mathbf{D}});\, c
    \big)
\right),
\label{eq:latent_loss}
\end{equation}

This produces paired data $(\tilde{\mathbf{D}}, \mathbf{D}_{gt})$ for text-conditioned training,
and $(\tilde{\mathbf{D}}^{\tau}, \mathbf{D}_{gt})$ for trajectory-conditioned training.
After generation, the predicted grid $\hat{\mathbf{D}}$ is split into $9$ frames for quantitative evaluation
and qualitative visualization.

\section{Experiments}
\subsection{Setup and Implementation}
\paragraph{Datasets:} JacoPlay, Bridge, RT1 dataset
We evaluate on three real-world manipulation datasets. 

\noindent \textbf{JacoPlay} provides 1{,}085 teleoperated episodes on a Kinova Jaco arm with clear observations, actions, and natural-language goal annotations; we use the official release for training and evaluation \cite{dass2023jacoplay}. 

\noindent \textbf{BridgeData~V2} is a large, diverse collection of real robot manipulation trajectories designed for scalable robot learning, comprising 53{,}896 trajectories across 24 environments; we follow the public CoRL/PMLR release and its splits. We randomly choose 9{,}967 samples for training and 515 for test \cite{pmlr-v229-walke23a}.

\noindent \textbf{RT-1} accompanies the Robotics Transformer study with a real-world dataset of $\sim$130k episodes spanning 700+ tasks collected by a fleet of 13 robots over 17 months; we cite the RSS’23 publication and project resources. We randomly choose 9{,}067 samples for training and 698 for test \cite{Brohan-RSS-23}.

\subsection{Quantitative Results}
\begin{figure*}[t]
    \centering
    \includegraphics[width=\textwidth]{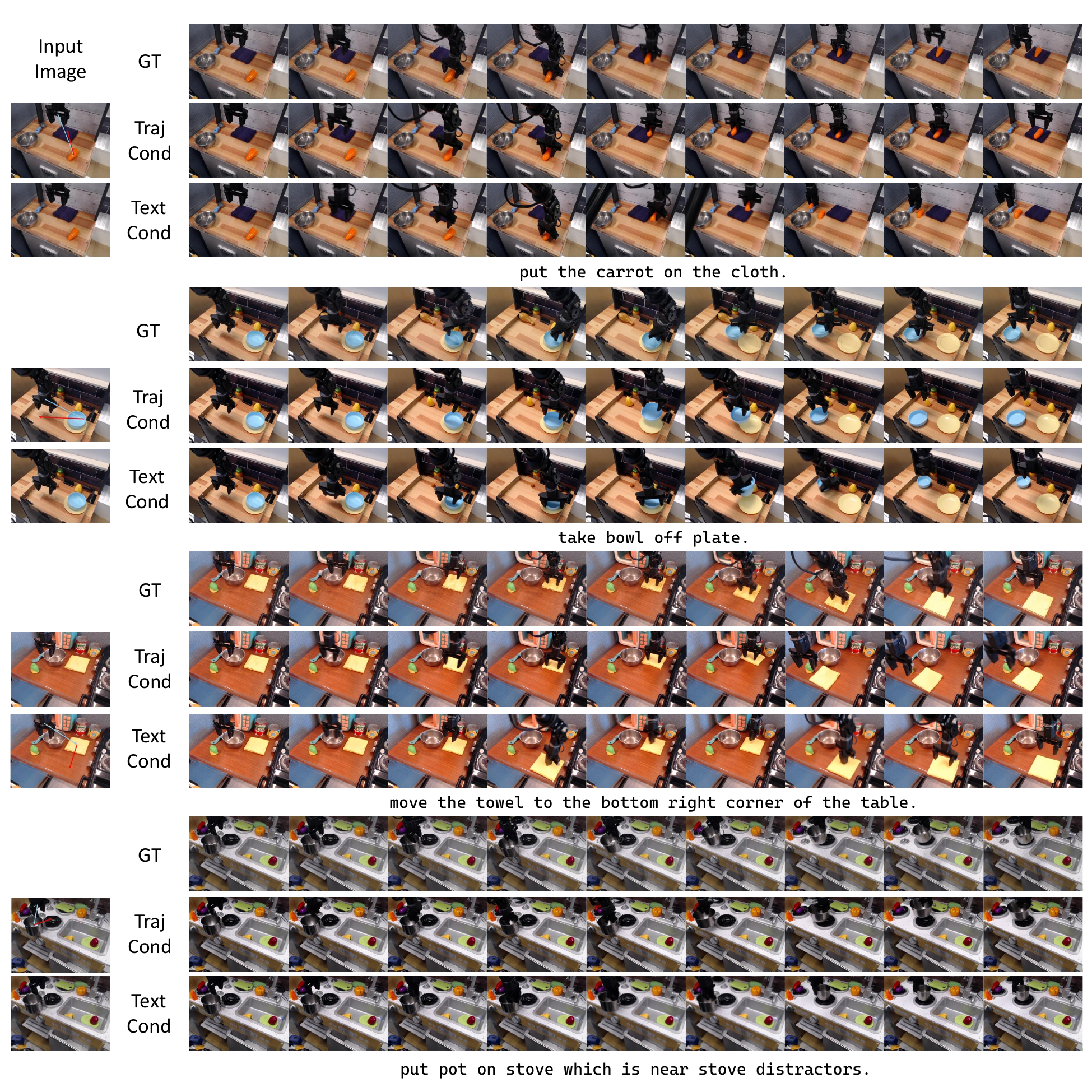}
    \caption{\textbf{Qualitative comparisons between text-conditioned and trajectory-conditioned generation.}
    Each row shows a 9-frame sequence arranged in temporal order.
    The top example (\textit{put the carrot on the cloth}) and bottom example (\textit{take bowl off plate})
    illustrate how both conditioning strategies interpret the same initial frame differently.
    The text-conditioned model relies solely on semantic understanding of the prompt to identify
    the correct object and its intended motion,
    while the trajectory-conditioned model follows the spatial path provided by the overlaid end-effector trace.
    }
    \label{fig:visual_comparisons}
\end{figure*}
\paragraph{Evaluation Metrics.}
We report four standard metrics for video generation and robotic control quality.

\textbf{Fr\'echet Video Distance (FVD).}
FVD measures the Wasserstein-2 (Fr\'echet) distance between multivariate Gaussians fitted to deep video features of real and generated clips (typically I3D features), capturing both spatial fidelity and temporal coherence \cite{Unterthiner2018FVD,Unterthiner2019FVDOpenReview}.
Let $(\mu_r,\Sigma_r)$ and $(\mu_g,\Sigma_g)$ be the empirical mean and covariance of real and generated video features, respectively.
\begin{equation}
\mathrm{FVD} = \|\mu_r - \mu_g\|_2^2
+ \mathrm{Tr}\!\Big(
    \Sigma_r + \Sigma_g - 2(\Sigma_r^{1/2}\Sigma_g\,\Sigma_r^{1/2})^{1/2}
\Big),
\label{eq:fvd}
\end{equation}

Lower is better. While widely used, recent analyses note potential content biases; thus we also report frame-wise metrics for completeness.

\textbf{Structural Similarity (SSIM).}
SSIM assesses perceptual similarity by comparing luminance $l$, contrast $c$, and structure $s$ between a generated frame $x$ and its reference $y$ \cite{wang2004ssim}:
\begin{equation}
\mathrm{SSIM}(x, y) =
\frac{(2\mu_x\mu_y + C_1)(2\sigma_{xy} + C_2)}
     {(\mu_x^2 + \mu_y^2 + C_1)(\sigma_x^2 + \sigma_y^2 + C_2)}\,,
\label{eq:ssim}
\end{equation}

where $\mu$ and $\sigma$ are local means and variances, $\sigma_{xy}$ is the covariance, and $C_1,C_2$ are stabilizers. Higher is better.

\textbf{Mean Squared Error (MSE).}
MSE is the average squared pixel error between a generated frame $x$ and reference $y$:
\begin{equation}
\mathrm{MSE}(x, y) =
\frac{1}{N} \sum_{i=1}^{N} \big(x_i - y_i\big)^2,
\label{eq:mse}
\end{equation}
where $N$ denotes the total number of pixels, and $x_i$, $y_i$ are the intensity values of the $i$-th pixel in the generated and reference images, respectively.
It serves as a classical full-reference fidelity metric complementary to SSIM \cite{wang2004ssim}. Lower is better.

\textbf{Success Rate.}
Following robotics evaluation practice (e.g., RT-1), we compute the proportion of evaluation episodes whose terminal state satisfies a task-specific success predicate defined by the dataset protocol or instruction (e.g., object placed in target receptacle) \cite{Brohan2023RT1}. Formally, for $K$ trials with binary outcomes $s_k\!\in\!\{0,1\}$,

\begin{equation}
\mathrm{SuccessRate} = \frac{1}{K} \sum_{k=1}^{K} s_k\,,
\label{eq:success_rate}
\end{equation}

\paragraph{Main Results.}
We evaluate the two experimental branches—\emph{text-conditioned generation} (baseline)
and \emph{trajectory-conditioned generation} (our proposed spatial control variant)—
across the JacoPlay, BridgeV2, and RT-1 datasets using four complementary metrics mentioned above.

\begin{table}[htbp]
\centering
\caption{\textbf{Main results across datasets.} Mean over 100 episodes (3 seeds).
FVD↓, SSIM↑, MSE↓, Success↑.}
\vspace{3pt}

\resizebox{\columnwidth}{!}{%
\begin{tabular}{llcccc}
\toprule
\textbf{Dataset} & \textbf{Method} & \textbf{FVD↓} & \textbf{SSIM↑} & \textbf{MSE↓} & \textbf{Success↑} \\
\midrule
\multirow{2}{*}{JacoPlay}
 & \textbf{Text} & 490.7 & 0.797 & 0.00695 & 80.6\% \\
 & \textbf{Traj} & 503.37 & 0.802 & 0.00680 & 74.0\% \\
\midrule
\multirow{2}{*}{BridgeV2}
 & \textbf{Text} & 644.2 & 0.733 & 0.0135 & 73.2\% \\
 & \textbf{Traj} & 693.2 & 0.726 & 0.0152  & 70.9\% \\
\midrule
\multirow{2}{*}{RT-1}
 & \textbf{Text} & 698.0 & 0.727 & 0.0118 & 72.4\% \\
 & \textbf{Traj} & 688.1 & 0.731 & 0.0117 & 81.7\% \\
\bottomrule
\end{tabular}
} 

\label{tab:all_main}
\end{table}

\subsection{Qualitative Results}

We compare the two conditioning modes—\textit{text-conditioned} and \textit{trajectory-conditioned} generation—on representative manipulation tasks. 
As shown in Fig.~\ref{fig:visual_comparisons}, the text-conditioned model excels at semantic grounding: it identifies the correct target object and generates plausible interactions. 
In contrast, the trajectory-conditioned model emphasizes spatial precision, producing motions that closely follow the provided 2D end-effector path for fine-grained control. 
Both modes generate temporally coherent sequences with smooth and physically consistent transitions between actions. 
The grid-based layout naturally enforces frame-to-frame continuity via local attention, maintaining stable backgrounds and consistent object–arm dynamics across time. 
Together, these results show that text conditioning contributes semantic intent, while trajectory conditioning provides geometric accuracy.

\subsection{Ablation Studies}

\begin{figure}[htbp]
  \centering
  \setlength{\tabcolsep}{2pt}
  \renewcommand{\arraystretch}{0.9}
  \begin{tabular}{ccc}
    \includegraphics[width=0.32\columnwidth]{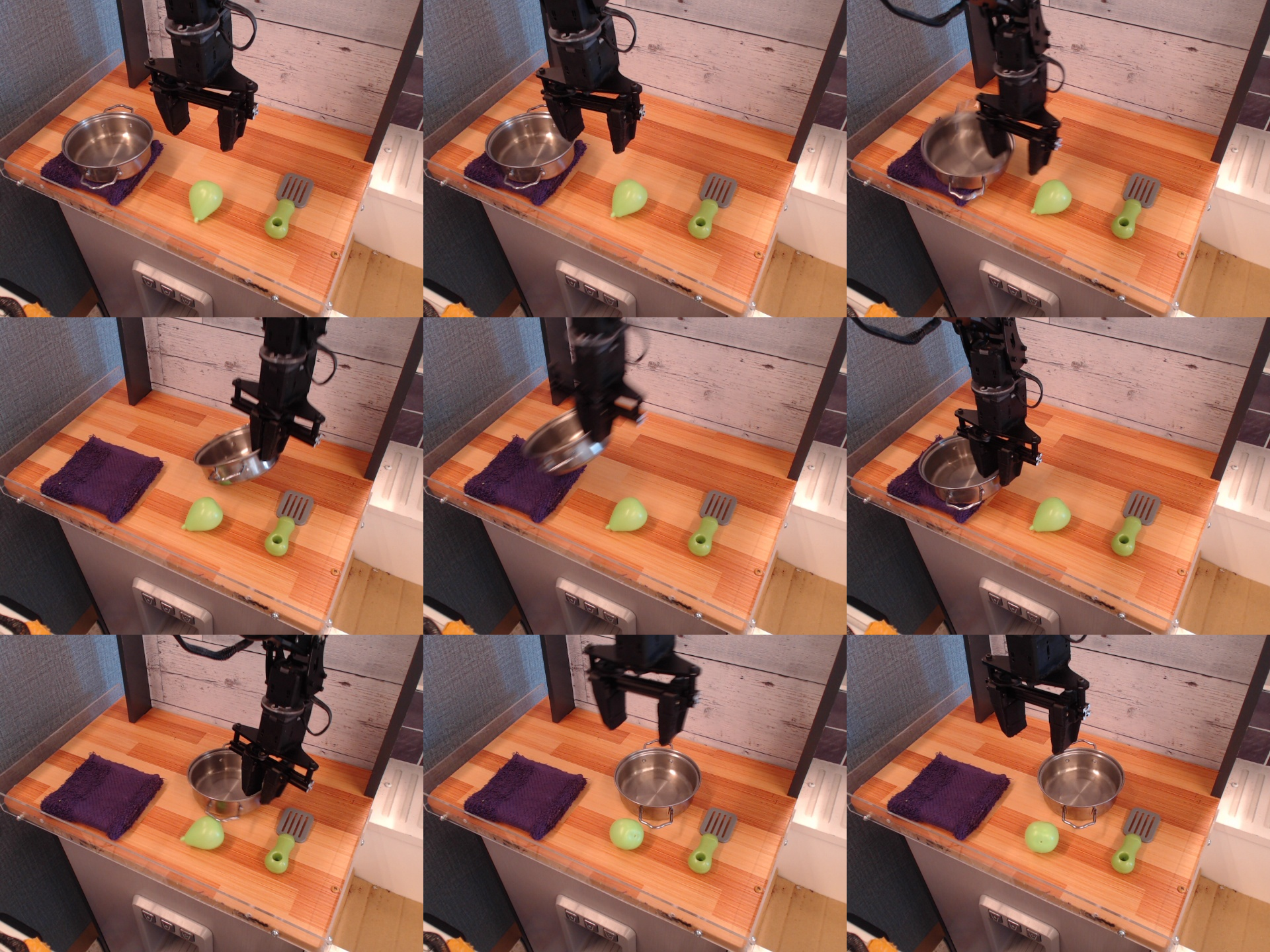} &
    \includegraphics[width=0.32\columnwidth]{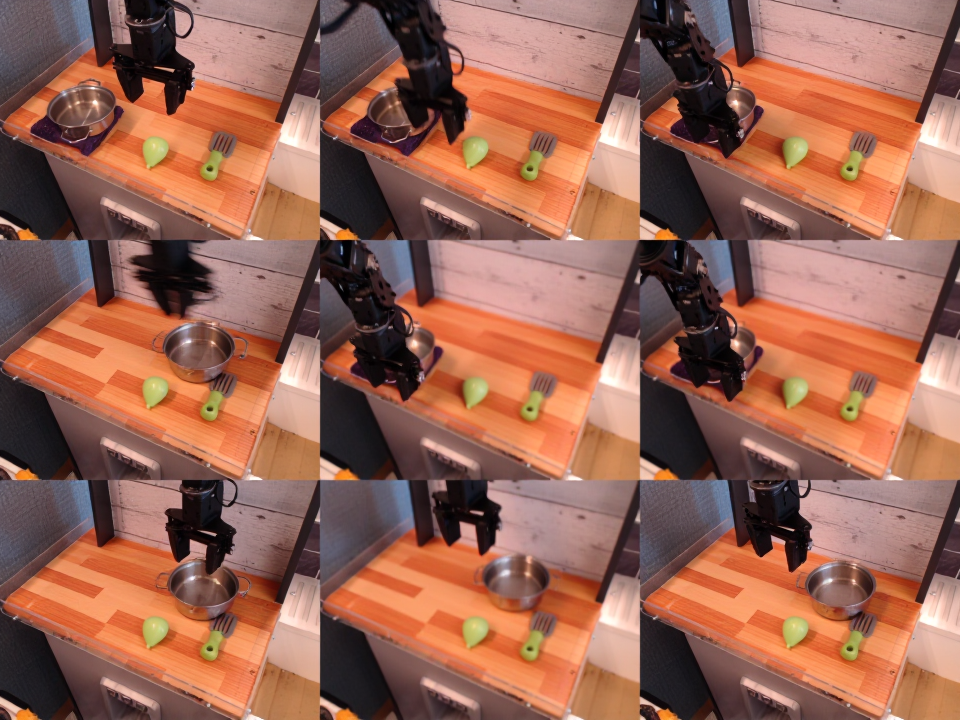} &
    \includegraphics[width=0.32\columnwidth]{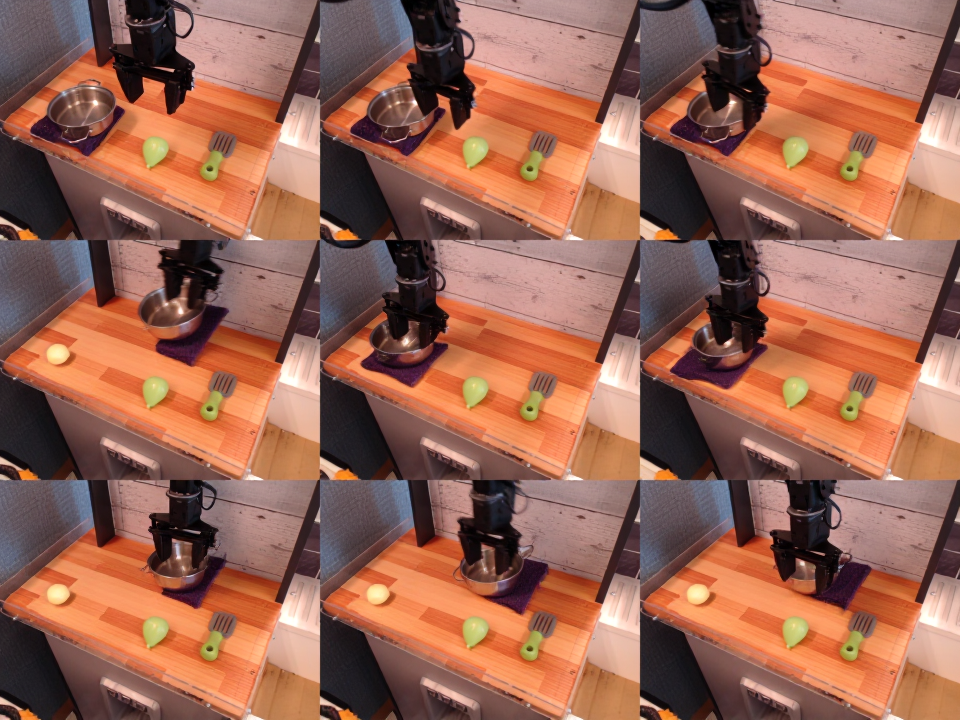} \\
    \small GT &
    \small Text-cond. &
    \small Traj-cond. \\[2pt]
    \includegraphics[width=0.32\columnwidth]{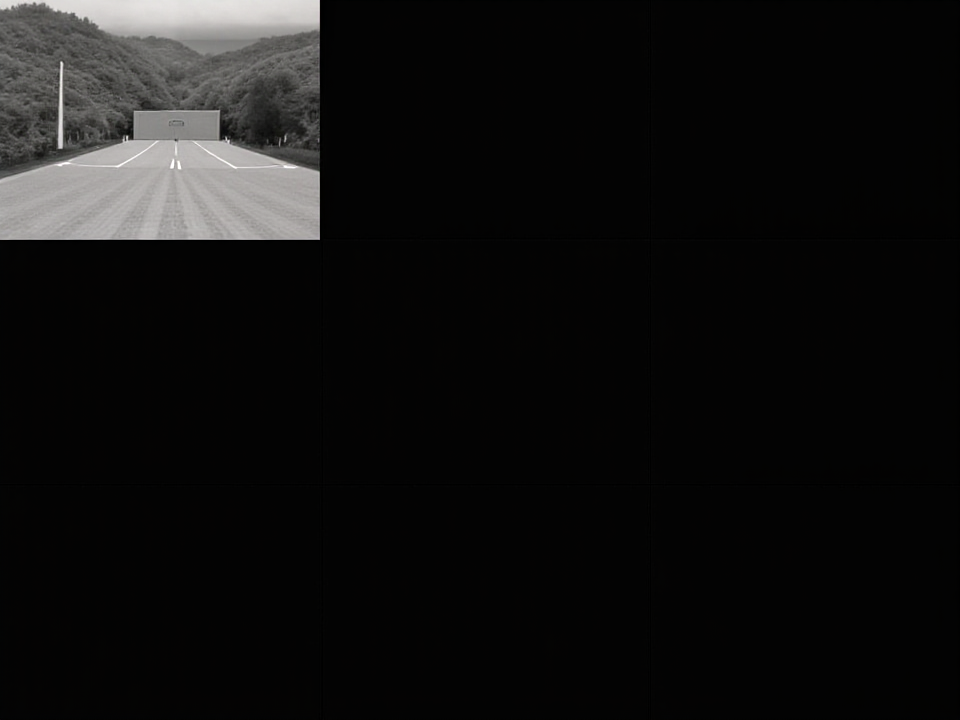} &
    \includegraphics[width=0.32\columnwidth]{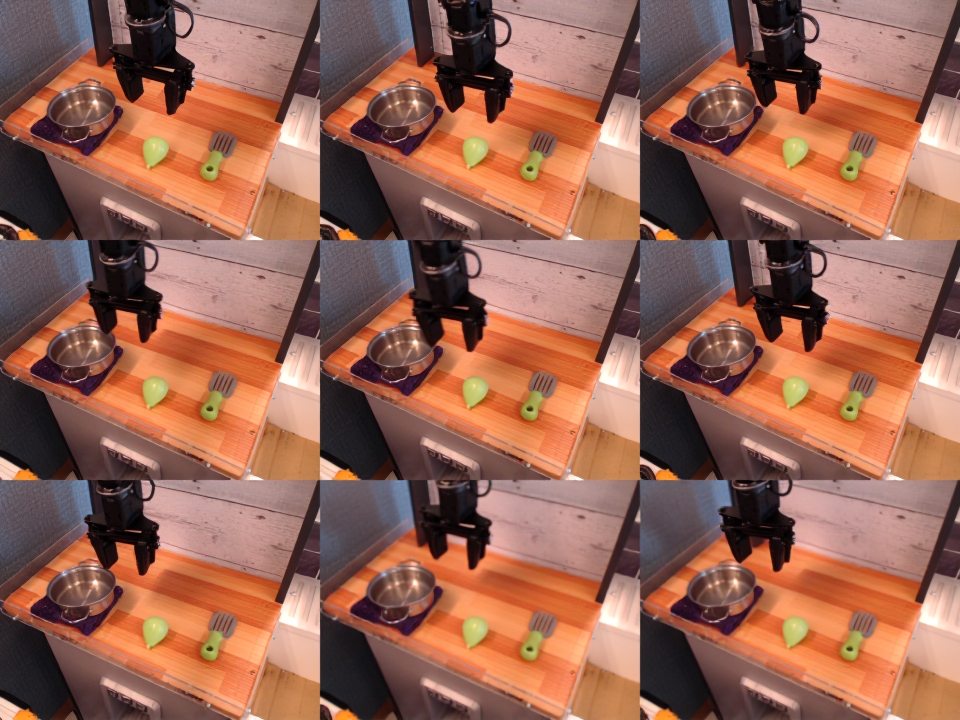} &
    \includegraphics[width=0.32\columnwidth]{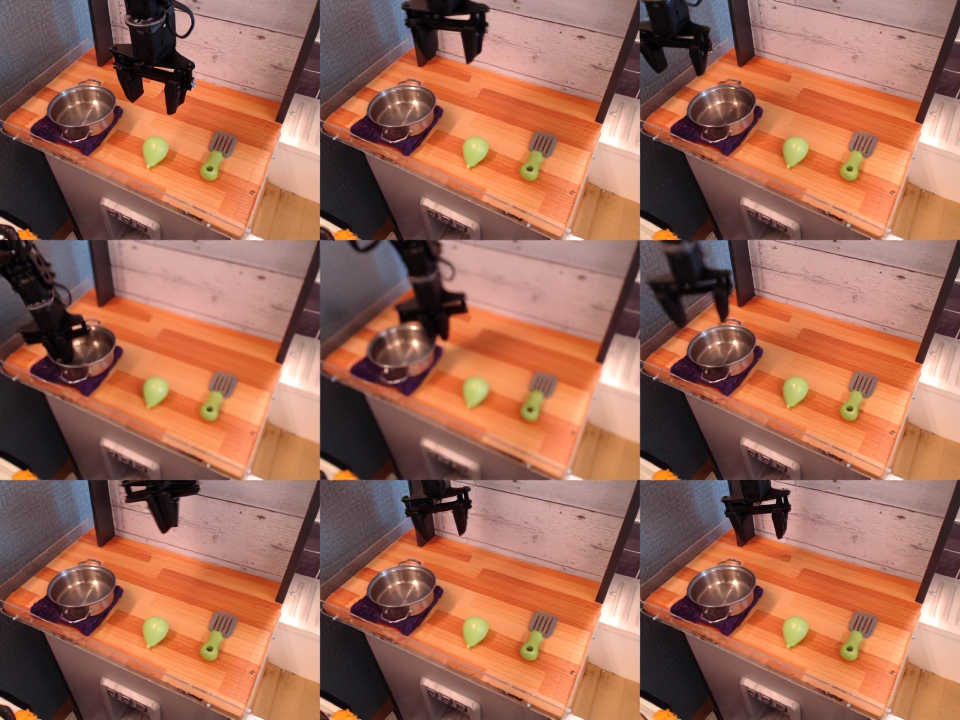} \\
    \small w/o LoRA &
    \small w/o Text &
    \small w/o Traj \\
  \end{tabular}
  \caption{\textbf{Ablation on the instruction “place the pot behind the green pear.”}
  Top: full models with LoRA and correct conditioning.  
  Bottom: removing LoRA or conditioning causes incoherent or aimless motions, while full models complete the instructed placement.}
  \label{fig:qual_ablation_grid}
\end{figure}

To isolate the contribution of each component, we perform ablations on BridgeV2 while keeping all training settings fixed. Fig \ref{fig:qual_ablation_grid} Table \ref{tab:ablation}

\paragraph{Effect of LoRA Adaptation.}
\textbf{w/o LoRA (frozen backbone)} tests whether the pretrained generator can synthesize manipulation sequences on its own.  
The frozen model fails to produce coherent grids, showing that general image priors alone do not transfer to robot-motion generation.  
LoRA is therefore essential for injecting task-specific motion and interaction semantics with minimal parameters.

\paragraph{Effect of Text Prompt Templates.}
Using raw dataset instructions (no template) noticeably degrades image quality and alignment.  
Without explicit cues about grid structure and manipulation intent, the arm often moves aimlessly and fails to pick the correct object, demonstrating that prompt structure meaningfully improves semantic grounding.

\paragraph{Effect of Trajectory Overlay.}
Removing rendered 2D trajectory from the trajectory-conditioned branch leads to clear failures: arm motions drift, objects barely move, and the model no longer executes the intended spatial path.  
This confirms that the visual trajectory overlay provides crucial spatial conditioning for precise, target-directed motion.

\begin{table}[t]
\centering
\caption{\textbf{Ablation results on BridgeV2.}
Each ablation modifies a single component from the full model.
↓ lower is better; ↑ higher is better.}
\vspace{3pt}
\resizebox{\columnwidth}{!}{%
\begin{tabular}{lcccc}
\toprule
\textbf{Configuration} & \textbf{FVD↓} & \textbf{SSIM↑} & \textbf{MSE↓} & \textbf{Success↑} \\
\midrule
Full model (Traj, grid)     & \textbf{644.2} & \textbf{0.733} & \textbf{0.0135} & \textbf{73.2\%} \\
Full model (Text, grid)     & \textbf{693.2} & \textbf{0.726} & \textbf{0.0152} & \textbf{70.9\%} \\
w/o LoRA (frozen backbone)       & 4377.1 & 0.064 & 0.1785 & 0\% \\
w/o Prompt Template (Text)            & 843.4 & 0.754 & 0.0153 & 2.5\% \\
w/o Trajectory Overlay (Traj)           & 720.0 & 0.749 & 0.0157 & 3.9\% \\
\bottomrule
\end{tabular}
}
\label{tab:ablation}
\end{table}

\subsection{Failure Cases and Limitations}

Our method remains stable across diverse scenes, but we occasionally observe minor visual inconsistencies. 
Slight tone or texture variations may appear between grid tiles due to independent latent sampling, and in rare cases, mild misalignment can occur along tile boundaries. 
These effects are visually negligible and do not affect the semantic correctness or temporal coherence of the generated manipulation sequences.
\section{Conclusion}

We have presented a controllable robotic video synthesis framework that adapts a pretrained image generator into a video-like model through lightweight LoRA adaptation. By reformulating robot manipulation sequences as $3{\times}3$ grid images, our approach unifies text- and trajectory-conditioned generation within a single architecture, leveraging the compositional priors of large image models for temporal reasoning. Experiments on multiple robot datasets demonstrate that the model not only generates coherent and physically plausible motion but also achieves competitive quantitative performance under both semantic and spatial conditioning. 

Beyond these results, our study highlights a broader insight: high-capacity image generators, when properly conditioned and adapted, can serve as visual planners capable of synthesizing action-consistent manipulation sequences from static inputs. 
{
    \small
    \bibliographystyle{ieeenat_fullname}
    \bibliography{main}

@String(CVPR= {IEEE Conf. Comput. Vis. Pattern Recog.})

@String(ICCV= {Int. Conf. Comput. Vis.})

@String(ICLR = {Int. Conf. Learn. Represent.})

@String(AAAI = {AAAI})

@String(CVPR  = {CVPR})

@String(ICCV  = {ICCV})

@String(ICLR  = {ICLR})

@article{ma2025followfaster,
  title={Follow-your-emoji-faster: Towards efficient, fine-controllable, and expressive freestyle portrait animation},
  author={Ma, Yue and Yan, Zexuan and Liu, Hongyu and Wang, Hongfa and Pan, Heng and He, Yingqing and Yuan, Junkun and Zeng, Ailing and Cai, Chengfei and Shum, Heung-Yeung and others},
  journal={arXiv preprint arXiv:2509.16630},
  year={2025}
}

@article{ma2025controllable,
  title={Controllable Video Generation: A Survey},
  author={Ma, Yue and Feng, Kunyu and Hu, Zhongyuan and Wang, Xinyu and Wang, Yucheng and Zheng, Mingzhe and He, Xuanhua and Zhu, Chenyang and Liu, Hongyu and He, Yingqing and others},
  journal={arXiv preprint arXiv:2507.16869},
  year={2025}
}

@article{ma2025followcreation,
  title={Follow-Your-Creation: Empowering 4D Creation through Video Inpainting},
  author={Ma, Yue and Feng, Kunyu and Zhang, Xinhua and Liu, Hongyu and Zhang, David Junhao and Xing, Jinbo and Zhang, Yinhan and Yang, Ayden and Wang, Zeyu and Chen, Qifeng},
  journal={arXiv preprint arXiv:2506.04590},
  year={2025}
}

@article{ma2025followyourmotion,
  title={Follow-Your-Motion: Video Motion Transfer via Efficient Spatial-Temporal Decoupled Finetuning},
  author={Ma, Yue and Liu, Yulong and Zhu, Qiyuan and Yang, Ayden and Feng, Kunyu and Zhang, Xinhua and Li, Zhifeng and Han, Sirui and Qi, Chenyang and Chen, Qifeng},
  journal={arXiv preprint arXiv:2506.05207},
  year={2025}
}

@inproceedings{ma2024followyouremoji,
  title={Follow-your-emoji: Fine-controllable and expressive freestyle portrait animation},
  author={Ma, Yue and Liu, Hongyu and Wang, Hongfa and Pan, Heng and He, Yingqing and Yuan, Junkun and Zeng, Ailing and Cai, Chengfei and Shum, Heung-Yeung and Liu, Wei and others},
  booktitle={SIGGRAPH Asia 2024 Conference Papers},
  pages={1--12},
  year={2024}
}

@inproceedings{ma2025followyourclick,
  title={Follow-Your-Click: Open-domain Regional Image Animation via Motion Prompts},
  author={Ma, Yue and He, Yingqing and Wang, Hongfa and Wang, Andong and Shen, Leqi and Qi, Chenyang and Ying, Jixuan and Cai, Chengfei and Li, Zhifeng and Shum, Heung-Yeung and others},
  booktitle={Proceedings of the AAAI Conference on Artificial Intelligence},
  volume={39},
  number={6},
  pages={6018--6026},
  year={2025}
}

@inproceedings{peebles2023dit,
  title     = {Scalable Diffusion Models with Transformers},
  author    = {Peebles, William and Xie, Saining},
  booktitle = {Proceedings of the IEEE/CVF International Conference on Computer Vision (ICCV)},
  year      = {2023},
  pages     = {4195--4205},
  url       = {https://openaccess.thecvf.com/content/ICCV2023/html/Peebles_Scalable_Diffusion_Models_with_Transformers_ICCV_2023_paper.html},
  eprint    = {2212.09748},
  archivePrefix = {arXiv}
}

@article{zhang2023controlnet,
  title   = {Adding Conditional Control to Text-to-Image Diffusion Models},
  author  = {Zhang, Lvmin and Rao, Anyi and Agrawala, Maneesh},
  journal = {arXiv preprint arXiv:2302.05543},
  year    = {2023},
  url     = {https://arxiv.org/abs/2302.05543}
}

@article{bartal2024lumiere,
  title   = {Lumiere: A Space-Time Diffusion Model for Video Generation},
  author  = {Bar-Tal, Omer and Chefer, Hila and Tov, Omer and Herrmann, Charles and Paiss, Roni and Zada, Shiran and Ephrat, Ariel and Hur, Junhwa and Liu, Guanghui and Raj, Amit and Li, Yuanzhen and Rubinstein, Michael and Michaeli, Tomer and Wang, Oliver and Sun, Deqing and Dekel, Tali and Mosseri, Inbar},
  journal = {arXiv preprint arXiv:2401.12945},
  year    = {2024},
  url     = {https://arxiv.org/abs/2401.12945},
  note    = {A SIGGRAPH Asia 2024 Technical Communications version is available},
  doi     = {10.1145/3680528.3687614}
}

@misc{openai2024sora,
  title        = {Video generation models as world simulators},
  author       = {{OpenAI}},
  howpublished = {\url{https://openai.com/index/video-generation-models-as-world-simulators/}},
  year         = {2024},
  note         = {Accessed: 2025-10-28}
}

@article{melnik2024_videodiff_survey,
  title   = {Video Diffusion Models: A Survey},
  author  = {Melnik, Andrew and Ljubljanac, Michal and Lu, Cong and Yan, Qi and Ren, Weiming and Ritter, Helge},
  journal = {arXiv preprint arXiv:2405.03150},
  year    = {2024},
  url     = {https://arxiv.org/abs/2405.03150}
}

@inproceedings{li2025trackdiffusion,
  title     = {TrackDiffusion: Tracklet-Conditioned Video Generation via Diffusion Models},
  author    = {Li, Pengxiang and Chen, Kai and Liu, Zhili and Gao, Ruiyuan and Hong, Lanqing and Yeung, Dit-Yan and Lu, Huchuan and Jia, Xu},
  booktitle = {Proceedings of the IEEE/CVF Winter Conference on Applications of Computer Vision (WACV)},
  year      = {2025},
  pages     = {3539--3548},
  url       = {https://openaccess.thecvf.com/content/WACV2025/html/Li_TrackDiffusion_Tracklet-Conditioned_Video_Generation_via_Diffusion_Models_WACV_2025_paper.html},
  eprint    = {2312.00651},
  archivePrefix = {arXiv}
}

@article{geng2025motionprompting,
  title   = {Controlling Video Generation with Motion Trajectories},
  author  = {Geng, Di and Lin, Yuwei and Xu, Danelljan and Meinhardt, Tim and Lu, Songhua and Danelljan, Martin and others},
  journal = {arXiv preprint arXiv:2412.02700},
  year    = {2025},
  url     = {https://arxiv.org/abs/2412.02700},
  note    = {CVPR 2025 (to appear)}
}

@article{hu2021lora,
  title   = {LoRA: Low-Rank Adaptation of Large Language Models},
  author  = {Hu, Edward J. and Shen, Yelong and Wallis, Phillip and Allen-Zhu, Zeyuan and Li, Yuanzhi and Wang, Shean and Wang, Lu and Chen, Weizhu},
  journal = {arXiv preprint arXiv:2106.09685},
  year    = {2021},
  url     = {https://arxiv.org/abs/2106.09685}
}

@article{patel2025rigvid,
  title   = {Robotic Manipulation by Imitating Generated Videos (RIGVid)},
  author  = {Patel, Shivansh and Mohan, Shraddhaa and Lazebnik, Svetlana and others},
  journal = {arXiv preprint arXiv:2507.00990},
  year    = {2025},
  url     = {https://arxiv.org/abs/2507.00990},
  note    = {Project page: \url{https://rigvid-robot.github.io/}}
}

@inproceedings{bharadhwaj2025gen2act,
  title     = {Gen2Act: Human Video Generation in Novel Scenarios Enables Generalizable Robot Manipulation},
  author    = {Bharadhwaj, Homanga and Dwibedi, Debidatta and Gupta, Abhinav and Tulsiani, Shubham and Doersch, Carl and Xiao, Ted and Shah, Dhruv and Xia, Fei and Sadigh, Dorsa and Kirmani, Sean},
  booktitle = {Proceedings of the Conference on Robot Learning (CoRL)},
  year      = {2025},
  url       = {https://openreview.net/pdf/71b0a6b0f5f4acafe22f89ad148d5d475e9dd2a2.pdf},
  eprint    = {2409.16283},
  archivePrefix = {arXiv},
  note      = {Project: \url{https://homangab.github.io/gen2act/}}
}

@inproceedings{walke2023bridgedatav2,
  title     = {BridgeData V2: A Dataset for Robot Learning at Scale},
  author    = {Walke, Homer Rich and Black, Kevin and Zhao, Tony Z. and Vuong, Quan and Zheng, Chongyi and Hansen-Estruch, Philippe and He, Andre Wang and Myers, Vivek and Kim, Moo Jin and Du, Max and Lee, Abraham and Fang, Kuan and Finn, Chelsea and Levine, Sergey},
  booktitle = {Proceedings of The 7th Conference on Robot Learning (CoRL), Proceedings of Machine Learning Research (PMLR)},
  year      = {2023},
  pages     = {1723--1736},
  url       = {https://proceedings.mlr.press/v229/walke23a/walke23a.pdf},
  eprint    = {2308.12952},
  archivePrefix = {arXiv}
}

@software{dass2023jacoplay,
  author  = {Dass, Shivin and Yapeter, Jullian and Zhang, Jesse and Zhang, Jiahui
             and Pertsch, Karl and Nikolaidis, Stefanos and Lim, Joseph J.},
  title   = {CLVR Jaco Play Dataset},
  url     = {https://github.com/clvrai/clvr_jaco_play_dataset},
  version = {1.0.0},
  year    = {2023}
}

@InProceedings{pmlr-v229-walke23a,
  title     = {BridgeData V2: A Dataset for Robot Learning at Scale},
  author    = {Walke, Homer Rich and Black, Kevin and Zhao, Tony Z.
               and Vuong, Quan and Zheng, Chongyi and Hansen-Estruch, Philippe
               and He, Andre Wang and Myers, Vivek and Kim, Moo Jin and Du, Max
               and Lee, Abraham and Fang, Kuan and Finn, Chelsea and Levine, Sergey},
  booktitle = {Proceedings of The 7th Conference on Robot Learning},
  pages     = {1723--1736},
  year      = {2023},
  editor    = {Tan, Jie and Toussaint, Marc and Darvish, Kourosh},
  volume    = {229},
  series    = {Proceedings of Machine Learning Research},
  month     = {06--09 Nov},
  publisher = {PMLR},
  url       = {https://proceedings.mlr.press/v229/walke23a.html}
}

@INPROCEEDINGS{Brohan-RSS-23,
  AUTHOR    = {Anthony Brohan AND Noah Brown AND Justice Carbajal AND Yevgen Chebotar
               AND Joseph Dabis AND Chelsea Finn AND Keerthana Gopalakrishnan
               AND Karol Hausman AND Alexander Herzog AND Jasmine Hsu AND Julian Ibarz
               AND Brian Ichter AND Alex Irpan AND Tomas Jackson AND Sally Jesmonth
               AND Nikhil Joshi AND Ryan Julian AND Dmitry Kalashnikov AND Yuheng Kuang
               AND Isabel Leal AND Kuang-Huei Lee AND Sergey Levine AND Yao Lu
               AND Utsav Malla AND Deeksha Manjunath AND Igor Mordatch AND Ofir Nachum
               AND Carolina Parada AND Jodilyn Peralta AND Emily Perez AND Karl Pertsch
               AND Jornell Quiambao AND Kanishka Rao AND Michael S Ryoo AND Grecia Salazar
               AND Pannag R Sanketi AND Kevin Sayed AND Jaspiar Singh AND Sumedh Sontakke
               AND Austin Stone AND Clayton Tan AND Huong Tran AND Vincent Vanhoucke
               AND Steve Vega AND Quan H Vuong AND Fei Xia AND Ted Xiao AND Peng Xu
               AND Sichun Xu AND Tianhe Yu AND Brianna Zitkovich},
  TITLE     = {{RT-1: Robotics Transformer for Real-World Control at Scale}},
  BOOKTITLE = {Proceedings of Robotics: Science and Systems},
  YEAR      = {2023},
  ADDRESS   = {Daegu, Republic of Korea},
  MONTH     = {July},
  DOI       = {10.15607/RSS.2023.XIX.025}
}

@article{Unterthiner2018FVD,
  title   = {Towards Accurate Generative Models of Video: A New Metric \& Challenges},
  author  = {Unterthiner, Thomas and van Steenkiste, Sjoerd and Kurach, Karol and Marinier, Rapha{\"e}l and Michalski, Marcin and Gelly, Sylvain},
  journal = {arXiv},
  eprint  = {1812.01717},
  year    = {2018},
  url     = {https://arxiv.org/abs/1812.01717}
}

@inproceedings{Unterthiner2019FVDOpenReview,
  title     = {FVD: A New Metric for Video Generation},
  author    = {Unterthiner, Thomas and van Steenkiste, Sjoerd and Kurach, Karol and Marinier, Rapha{\"e}l and Michalski, Marcin and Gelly, Sylvain},
  booktitle = {Deep Generative Models for Highly Structured Data (DGS) @ ICLR},
  year      = {2019},
  url       = {https://openreview.net/forum?id=rylgEULtdN}
}

@article{Wang2004SSIM,
  author    = {Wang, Zhou and Bovik, Alan C. and Sheikh, Hamid R. and Simoncelli, Eero P.},
  title     = {Image Quality Assessment: From Error Visibility to Structural Similarity},
  journal   = {IEEE Transactions on Image Processing},
  volume    = {13},
  number    = {4},
  pages     = {600--612},
  year      = {2004},
  doi       = {10.1109/TIP.2003.819861},
  url       = {https://ieeexplore.ieee.org/document/1284395}
}

@inproceedings{Brohan2023RT1,
  author    = {Brohan, Anthony and Brown, Noah and Carbajal, Justice and Chebotar, Yevgen and Dabis, Joseph and Finn, Chelsea and Gopalakrishnan, Keerthana and Hausman, Karol and Herzog, Alexander and Hsu, Jasmine and Ibarz, Julian and Ichter, Brian and Irpan, Alex and Jackson, Tomas and Jesmonth, Sally and Joshi, Nikhil and Julian, Ryan and Kalashnikov, Dmitry and Kuang, Yuheng and Leal, Isabel and Lee, Kuang-Huei and Levine, Sergey and Lu, Yao and Malla, Utsav and Manjunath, Deeksha and Mordatch, Igor and Nachum, Ofir and Parada, Carolina and Peralta, Jodilyn and Perez, Emily and Pertsch, Karl and Quiambao, Jornell and Rao, Kanishka and Ryoo, Michael S. and Salazar, Grecia and Sanketi, Pannag R. and Sayed, Kevin and Singh, Jaspiar and Sontakke, Sumedh and Stone, Austin and Tan, Clayton and Tran, Huong and Vanhoucke, Vincent and Vega, Steve and Vuong, Quan H. and Xia, Fei and Xiao, Ted and Xu, Peng and Xu, Sichun and Yu, Tianhe and Zitkovich, Brianna},
  title     = {RT-1: Robotics Transformer for Real-World Control at Scale},
  booktitle = {Proceedings of Robotics: Science and Systems (RSS)},
  address   = {Daegu, Republic of Korea},
  month     = {July},
  year      = {2023},
  doi       = {10.15607/RSS.2023.XIX.025},
  url       = {https://www.roboticsproceedings.org/rss19/p025.pdf}
}

@article{cao2024_controllable_t2i_survey,
  title   = {Controllable Generation with Text-to-Image Diffusion Models: A Survey},
  author  = {Cao, Puyang and Zhou, Yang and others},
  journal = {arXiv preprint arXiv:2403.04279},
  year    = {2024},
  url     = {https://arxiv.org/abs/2403.04279}
}

@article{zhang2025easycontrol,
  title={Easycontrol: Adding efficient and flexible control for diffusion transformer},
  author={Zhang, Yuxuan and Yuan, Yirui and Song, Yiren and Wang, Haofan and Liu, Jiaming},
  journal={arXiv preprint arXiv:2503.07027},
  year={2025}
}

@inproceedings{zhang2024ssr,
  title={Ssr-encoder: Encoding selective subject representation for subject-driven generation},
  author={Zhang, Yuxuan and Song, Yiren and Liu, Jiaming and Wang, Rui and Yu, Jinpeng and Tang, Hao and Li, Huaxia and Tang, Xu and Hu, Yao and Pan, Han and others},
  booktitle={Proceedings of the IEEE/CVF Conference on Computer Vision and Pattern Recognition},
  pages={8069--8078},
  year={2024}
}

@article{song2025layertracer,
  title={LayerTracer: Cognitive-Aligned Layered SVG Synthesis via Diffusion Transformer},
  author={Song, Yiren and Chen, Danze and Shou, Mike Zheng},
  journal={arXiv preprint arXiv:2502.01105},
  year={2025}
}

@article{song2025makeanything,
  title={MakeAnything: Harnessing Diffusion Transformers for Multi-Domain Procedural Sequence Generation},
  author={Song, Yiren and Liu, Cheng and Shou, Mike Zheng},
  journal={arXiv preprint arXiv:2502.01572},
  year={2025}
}

@article{huang2025photodoodle,
  title={Photodoodle: Learning artistic image editing from few-shot pairwise data},
  author={Huang, Shijie and Song, Yiren and Zhang, Yuxuan and Guo, Hailong and Wang, Xueyin and Shou, Mike Zheng and Liu, Jiaming},
  journal={arXiv preprint arXiv:2502.14397},
  year={2025}
}

@article{guo2025any2anytryon,
  title={Any2AnyTryon: Leveraging Adaptive Position Embeddings for Versatile Virtual Clothing Tasks},
  author={Guo, Hailong and Zeng, Bohan and Song, Yiren and Zhang, Wentao and Zhang, Chuang and Liu, Jiaming},
  journal={arXiv preprint arXiv:2501.15891},
  year={2025}
}

@article{chen2025transanimate,
  title={Transanimate: Taming layer diffusion to generate rgba video},
  author={Chen, Xuewei and Chen, Zhimin and Song, Yiren},
  journal={arXiv preprint arXiv:2503.17934},
  year={2025}
}

@article{song2024processpainter,
  title={Processpainter: Learn painting process from sequence data},
  author={Song, Yiren and Huang, Shijie and Yao, Chen and Ye, Xiaojun and Ci, Hai and Liu, Jiaming and Zhang, Yuxuan and Shou, Mike Zheng},
  journal={arXiv preprint arXiv:2406.06062},
  year={2024}
}

@inproceedings{zhang2025stable,
  title={Stable-hair: Real-world hair transfer via diffusion model},
  author={Zhang, Yuxuan and Zhang, Qing and Song, Yiren and Zhang, Jichao and Tang, Hao and Liu, Jiaming},
  booktitle={Proceedings of the AAAI Conference on Artificial Intelligence},
  volume={39},
  number={10},
  pages={10348--10356},
  year={2025}
}

@article{zhang2024stable,
  title={Stable-makeup: When real-world makeup transfer meets diffusion model},
  author={Zhang, Yuxuan and Wei, Lifu and Zhang, Qing and Song, Yiren and Liu, Jiaming and Li, Huaxia and Tang, Xu and Hu, Yao and Zhao, Haibo},
  journal={arXiv preprint arXiv:2403.07764},
  year={2024}
}

@article{wan2024grid,
  title={Grid: Visual layout generation},
  author={Wan, Cong and Luo, Xiangyang and Cai, Zijian and Song, Yiren and Zhao, Yunlong and Bai, Yifan and He, Yuhang and Gong, Yihong},
  journal={arXiv preprint arXiv:2412.10718},
  year={2024}
}

@article{wang2025diffdecompose,
  title={DiffDecompose: Layer-Wise Decomposition of Alpha-Composited Images via Diffusion Transformers},
  author={Wang, Zitong and Zhao, Hang and Zhou, Qianyu and Lu, Xuequan and Li, Xiangtai and Song, Yiren},
  journal={arXiv preprint arXiv:2505.21541},
  year={2025}
}

@article{gong2025relationadapter,
  title={RelationAdapter: Learning and Transferring Visual Relation with Diffusion Transformers},
  author={Gong, Yan and Song, Yiren and Li, Yicheng and Li, Chenglin and Zhang, Yin},
  journal={arXiv preprint arXiv:2506.02528},
  year={2025}
}

@article{jiang2025personalized,
  title={Personalized Vision via Visual In-Context Learning},
  author={Jiang, Yuxin and Gu, Yuchao and Song, Yiren and Tsang, Ivor and Shou, Mike Zheng},
  journal={arXiv preprint arXiv:2509.25172},
  year={2025}
}

@inproceedings{li2023gligen,
  title     = {GLIGEN: Open-Set Grounded Text-to-Image Generation},
  author    = {Li, Yuheng and Liu, Haotian and Wu, Qingyang and Mu, Fangzhou and Yang, Jianwei and Gao, Jianfeng and Li, Chunyuan and Lee, Yong Jae},
  booktitle = {Proceedings of the IEEE/CVF Conference on Computer Vision and Pattern Recognition (CVPR)},
  year      = {2023},
  url       = {https://openaccess.thecvf.com/content/CVPR2023/papers/Li_GLIGEN_Open-Set_Grounded_Text-to-Image_Generation_CVPR_2023_paper.pdf},
  eprint    = {2301.07093},
  archivePrefix = {arXiv}
}

@misc{rt1,
      title={RT-1: Robotics Transformer for Real-World Control at Scale}, 
      author={Anthony Brohan and Noah Brown and Justice Carbajal and Yevgen Chebotar and Joseph Dabis and Chelsea Finn and Keerthana Gopalakrishnan and Karol Hausman and Alex Herzog and Jasmine Hsu and Julian Ibarz and Brian Ichter and Alex Irpan and Tomas Jackson and Sally Jesmonth and Nikhil J Joshi and Ryan Julian and Dmitry Kalashnikov and Yuheng Kuang and Isabel Leal and Kuang-Huei Lee and Sergey Levine and Yao Lu and Utsav Malla and Deeksha Manjunath and Igor Mordatch and Ofir Nachum and Carolina Parada and Jodilyn Peralta and Emily Perez and Karl Pertsch and Jornell Quiambao and Kanishka Rao and Michael Ryoo and Grecia Salazar and Pannag Sanketi and Kevin Sayed and Jaspiar Singh and Sumedh Sontakke and Austin Stone and Clayton Tan and Huong Tran and Vincent Vanhoucke and Steve Vega and Quan Vuong and Fei Xia and Ted Xiao and Peng Xu and Sichun Xu and Tianhe Yu and Brianna Zitkovich},
      year={2023},
      eprint={2212.06817},
      archivePrefix={arXiv},
      primaryClass={cs.RO},
      url={https://arxiv.org/abs/2212.06817}, 
}

@misc{roboflamingo,
      title={Vision-Language Foundation Models as Effective Robot Imitators}, 
      author={Xinghang Li and Minghuan Liu and Hanbo Zhang and Cunjun Yu and Jie Xu and Hongtao Wu and Chilam Cheang and Ya Jing and Weinan Zhang and Huaping Liu and Hang Li and Tao Kong},
      year={2024},
      eprint={2311.01378},
      archivePrefix={arXiv},
      primaryClass={cs.RO},
      url={https://arxiv.org/abs/2311.01378}, 
}

@misc{leo,
      title={An Embodied Generalist Agent in 3D World}, 
      author={Jiangyong Huang and Silong Yong and Xiaojian Ma and Xiongkun Linghu and Puhao Li and Yan Wang and Qing Li and Song-Chun Zhu and Baoxiong Jia and Siyuan Huang},
      year={2024},
      eprint={2311.12871},
      archivePrefix={arXiv},
      primaryClass={cs.CV},
      url={https://arxiv.org/abs/2311.12871}, 
}

@misc{gr1,
      title={Unleashing Large-Scale Video Generative Pre-training for Visual Robot Manipulation}, 
      author={Hongtao Wu and Ya Jing and Chilam Cheang and Guangzeng Chen and Jiafeng Xu and Xinghang Li and Minghuan Liu and Hang Li and Tao Kong},
      year={2023},
      eprint={2312.13139},
      archivePrefix={arXiv},
      primaryClass={cs.RO},
      url={https://arxiv.org/abs/2312.13139}, 
}

@misc{octo,
      title={Octo: An Open-Source Generalist Robot Policy}, 
      author={Octo Model Team and Dibya Ghosh and Homer Walke and Karl Pertsch and Kevin Black and Oier Mees and Sudeep Dasari and Joey Hejna and Tobias Kreiman and Charles Xu and Jianlan Luo and You Liang Tan and Lawrence Yunliang Chen and Pannag Sanketi and Quan Vuong and Ted Xiao and Dorsa Sadigh and Chelsea Finn and Sergey Levine},
      year={2024},
      eprint={2405.12213},
      archivePrefix={arXiv},
      primaryClass={cs.RO},
      url={https://arxiv.org/abs/2405.12213}, 
}

@misc{openvla,
      title={OpenVLA: An Open-Source Vision-Language-Action Model}, 
      author={Moo Jin Kim and Karl Pertsch and Siddharth Karamcheti and Ted Xiao and Ashwin Balakrishna and Suraj Nair and Rafael Rafailov and Ethan Foster and Grace Lam and Pannag Sanketi and Quan Vuong and Thomas Kollar and Benjamin Burchfiel and Russ Tedrake and Dorsa Sadigh and Sergey Levine and Percy Liang and Chelsea Finn},
      year={2024},
      eprint={2406.09246},
      archivePrefix={arXiv},
      primaryClass={cs.RO},
      url={https://arxiv.org/abs/2406.09246}, 
}

@misc{pi0,
      title={$\pi_0$: A Vision-Language-Action Flow Model for General Robot Control}, 
      author={Kevin Black and Noah Brown and Danny Driess and Adnan Esmail and Michael Equi and Chelsea Finn and Niccolo Fusai and Lachy Groom and Karol Hausman and Brian Ichter and Szymon Jakubczak and Tim Jones and Liyiming Ke and Sergey Levine and Adrian Li-Bell and Mohith Mothukuri and Suraj Nair and Karl Pertsch and Lucy Xiaoyang Shi and James Tanner and Quan Vuong and Anna Walling and Haohuan Wang and Ury Zhilinsky},
      year={2024},
      eprint={2410.24164},
      archivePrefix={arXiv},
      primaryClass={cs.LG},
      url={https://arxiv.org/abs/2410.24164}, 
}

@misc{uniact,
      title={Universal Actions for Enhanced Embodied Foundation Models}, 
      author={Jinliang Zheng and Jianxiong Li and Dongxiu Liu and Yinan Zheng and Zhihao Wang and Zhonghong Ou and Yu Liu and Jingjing Liu and Ya-Qin Zhang and Xianyuan Zhan},
      year={2025},
      eprint={2501.10105},
      archivePrefix={arXiv},
      primaryClass={cs.RO},
      url={https://arxiv.org/abs/2501.10105}, 
}

@misc{tinyvla,
      title={TinyVLA: Towards Fast, Data-Efficient Vision-Language-Action Models for Robotic Manipulation}, 
      author={Junjie Wen and Yichen Zhu and Jinming Li and Minjie Zhu and Kun Wu and Zhiyuan Xu and Ning Liu and Ran Cheng and Chaomin Shen and Yaxin Peng and Feifei Feng and Jian Tang},
      year={2025},
      eprint={2409.12514},
      archivePrefix={arXiv},
      primaryClass={cs.RO},
      url={https://arxiv.org/abs/2409.12514}, 
}

@misc{gr00t,
      title={GR00T N1: An Open Foundation Model for Generalist Humanoid Robots}, 
      author={NVIDIA and : and Johan Bjorck and Fernando Castañeda and Nikita Cherniadev and Xingye Da and Runyu Ding and Linxi "Jim" Fan and Yu Fang and Dieter Fox and Fengyuan Hu and Spencer Huang and Joel Jang and Zhenyu Jiang and Jan Kautz and Kaushil Kundalia and Lawrence Lao and Zhiqi Li and Zongyu Lin and Kevin Lin and Guilin Liu and Edith Llontop and Loic Magne and Ajay Mandlekar and Avnish Narayan and Soroush Nasiriany and Scott Reed and You Liang Tan and Guanzhi Wang and Zu Wang and Jing Wang and Qi Wang and Jiannan Xiang and Yuqi Xie and Yinzhen Xu and Zhenjia Xu and Seonghyeon Ye and Zhiding Yu and Ao Zhang and Hao Zhang and Yizhou Zhao and Ruijie Zheng and Yuke Zhu},
      year={2025},
      eprint={2503.14734},
      archivePrefix={arXiv},
      primaryClass={cs.RO},
      url={https://arxiv.org/abs/2503.14734}, 
}

@misc{genie3,
  title     = {Genie 3: A New Frontier for World Models},
  author    = {Ball, Philip J. and Bauer, J. and Belletti, F. and others},
  year      = {2025},
}

@misc{moviegen,
      title={Movie Gen: A Cast of Media Foundation Models}, 
      author={Adam Polyak and Amit Zohar and Andrew Brown and Andros Tjandra and Animesh Sinha and Ann Lee and Apoorv Vyas and Bowen Shi and Chih-Yao Ma and Ching-Yao Chuang and David Yan and Dhruv Choudhary and Dingkang Wang and Geet Sethi and Guan Pang and Haoyu Ma and Ishan Misra and Ji Hou and Jialiang Wang and Kiran Jagadeesh and Kunpeng Li and Luxin Zhang and Mannat Singh and Mary Williamson and Matt Le and Matthew Yu and Mitesh Kumar Singh and Peizhao Zhang and Peter Vajda and Quentin Duval and Rohit Girdhar and Roshan Sumbaly and Sai Saketh Rambhatla and Sam Tsai and Samaneh Azadi and Samyak Datta and Sanyuan Chen and Sean Bell and Sharadh Ramaswamy and Shelly Sheynin and Siddharth Bhattacharya and Simran Motwani and Tao Xu and Tianhe Li and Tingbo Hou and Wei-Ning Hsu and Xi Yin and Xiaoliang Dai and Yaniv Taigman and Yaqiao Luo and Yen-Cheng Liu and Yi-Chiao Wu and Yue Zhao and Yuval Kirstain and Zecheng He and Zijian He and Albert Pumarola and Ali Thabet and Artsiom Sanakoyeu and Arun Mallya and Baishan Guo and Boris Araya and Breena Kerr and Carleigh Wood and Ce Liu and Cen Peng and Dimitry Vengertsev and Edgar Schonfeld and Elliot Blanchard and Felix Juefei-Xu and Fraylie Nord and Jeff Liang and John Hoffman and Jonas Kohler and Kaolin Fire and Karthik Sivakumar and Lawrence Chen and Licheng Yu and Luya Gao and Markos Georgopoulos and Rashel Moritz and Sara K. Sampson and Shikai Li and Simone Parmeggiani and Steve Fine and Tara Fowler and Vladan Petrovic and Yuming Du},
      year={2025},
      eprint={2410.13720},
      archivePrefix={arXiv},
      primaryClass={cs.CV},
      url={https://arxiv.org/abs/2410.13720}, 
}

@misc{sora,
  author       = {OpenAI},
  title        = {Sora},
  year         = {2024},
  howpublished = {\url{https://openai.com/sora/}},
  note         = {Accessed: 2025-11-08}
}

@misc{gen3,
  author       = {Runway},
  title        = {Gen 3},
  year         = {2024},
  howpublished = {\url{https://runwayml.com/research/introducing-gen-3-alpha}},
  note         = {Accessed: 2025-11-08}
}

@misc{minimax,
  author       = {Minimax},
  title        = {Hailuo},
  year         = {2024},
  howpublished = {\url{https://hailuoai.com/video}},
  note         = {Accessed: 2025-11-08}
}

@misc{veo3,
  author       = {Google Deepmind},
  title        = {Veo 3},
  year         = {2025},
  howpublished = {\url{https://deepmind.google/technologies/veo/veo-3/}},
  note         = {Accessed: 2025-11-08}
}

@misc{hunyuanvideo,
      title={HunyuanVideo: A Systematic Framework For Large Video Generative Models}, 
      author={Weijie Kong and Qi Tian and Zijian Zhang and Rox Min and Zuozhuo Dai and Jin Zhou and Jiangfeng Xiong and Xin Li and Bo Wu and Jianwei Zhang and Kathrina Wu and Qin Lin and Junkun Yuan and Yanxin Long and Aladdin Wang and Andong Wang and Changlin Li and Duojun Huang and Fang Yang and Hao Tan and Hongmei Wang and Jacob Song and Jiawang Bai and Jianbing Wu and Jinbao Xue and Joey Wang and Kai Wang and Mengyang Liu and Pengyu Li and Shuai Li and Weiyan Wang and Wenqing Yu and Xinchi Deng and Yang Li and Yi Chen and Yutao Cui and Yuanbo Peng and Zhentao Yu and Zhiyu He and Zhiyong Xu and Zixiang Zhou and Zunnan Xu and Yangyu Tao and Qinglin Lu and Songtao Liu and Dax Zhou and Hongfa Wang and Yong Yang and Di Wang and Yuhong Liu and Jie Jiang and Caesar Zhong},
      year={2025},
      eprint={2412.03603},
      archivePrefix={arXiv},
      primaryClass={cs.CV},
      url={https://arxiv.org/abs/2412.03603}, 
}

@misc{i2vgenxl,
      title={I2VGen-XL: High-Quality Image-to-Video Synthesis via Cascaded Diffusion Models}, 
      author={Shiwei Zhang and Jiayu Wang and Yingya Zhang and Kang Zhao and Hangjie Yuan and Zhiwu Qin and Xiang Wang and Deli Zhao and Jingren Zhou},
      year={2023},
      eprint={2311.04145},
      archivePrefix={arXiv},
      primaryClass={cs.CV},
      url={https://arxiv.org/abs/2311.04145}, 
}

@misc{pyramidflow,
      title={Pyramidal Flow Matching for Efficient Video Generative Modeling}, 
      author={Yang Jin and Zhicheng Sun and Ningyuan Li and Kun Xu and Kun Xu and Hao Jiang and Nan Zhuang and Quzhe Huang and Yang Song and Yadong Mu and Zhouchen Lin},
      year={2025},
      eprint={2410.05954},
      archivePrefix={arXiv},
      primaryClass={cs.CV},
      url={https://arxiv.org/abs/2410.05954}, 
}
}


\end{document}